\documentclass[conference]{IEEEtran}
\IEEEoverridecommandlockouts
\usepackage{cite}
\usepackage{amsmath,amssymb,amsfonts}
\usepackage{algorithmic}
\usepackage{graphicx}
\usepackage{textcomp}
\usepackage{xcolor}
\def\BibTeX{{\rm B\kern-.05em{\sc i\kern-.025em b}\kern-.08em
    T\kern-.1667em\lower.7ex\hbox{E}\kern-.125emX}}
\begin{document}

\title{Guided Diffusion for the Extension of Machine Vision to Human Visual Perception}

\author{Takahiro Shindo \qquad Yui Tatsumi \qquad Taiju Watanabe \qquad Hiroshi Watanabe\\
Waseda University}
\maketitle

\begin{abstract}
   Image compression technology eliminates redundant information to enable efficient transmission and storage of images, serving both machine vision and human visual perception.
   For years, image coding focused on human perception has been well-studied, leading to the development of various image compression standards.
   On the other hand, with the rapid advancements in image recognition models, image compression for AI tasks, known as Image Coding for Machines (ICM), has gained significant importance.
   Therefore, scalable image coding techniques that address the needs of both machines and humans have become a key area of interest.
   Additionally, there is increasing demand for research applying the diffusion model, which can generate human-viewable images from a small amount of data to image compression methods for human vision.
   Image compression methods that use diffusion models can partially reconstruct the target image by guiding the generation process with a small amount of conditioning information.
   Inspired by the diffusion model's potential, we propose a method for extending machine vision to human visual perception using guided diffusion.
   Utilizing the diffusion model guided by the output of the ICM method, we generate images for human perception from random noise.
   Guided diffusion acts as a bridge between machine vision and human vision, enabling transitions between them without any additional bitrate overhead.
   The generated images then evaluated based on bitrate and image quality, and we compare their compression performance with other scalable image coding methods for humans and machines.
\end{abstract}

\begin{IEEEkeywords}
Image Coding for Machines, Scalable Image Coding, Guided Diffusion
\end{IEEEkeywords}

\section{Introduction}
\label{sec:intro}

Humans and machines process the information in images to support their respective vision.
Image compression is a technology designed to identify the critical information needed for these visions and to represent images using the minimal amount of data necessary.
This technology is vital for efficient image transmission and storage.
In recent years, the rapid advancements in deep learning have increased the demand for image compression methods tailored not only for human perception, as traditionally emphasized, but also for image recognition.
This shift is driven by the growing use of models for image recognition tasks like object detection and segmentation.
To address this need, the field of Image Coding for Machines (ICM) has emerged as an important area of research \cite{r1,r2,r3,r4}.

Studies have shown that low-frequency components in images play a significant role in human perception, which is why high-frequency components are often discarded during image compression.
For example, methods like JPEG \cite{s1} and JPEG2000 \cite{s2} use Discrete Cosine Transform (DCT) and Discrete Wavelet Transform (DWT), respectively, to process images in the frequency domain and eliminate unnecessary frequency components.
This approach reduces the amount of data required to represent the images.
In contrast, research has indicated that object regions and edges within images are critical for image recognition.
Consequently, machine-oriented image compression methods have been developed that discard irrelevant information while preserving these key features in the image \cite{s3,s4,s5,s6}.
These differences result in distinct decoded images depending on whether they are optimized for human or machine vision.
Specifically, decoded images intended for machine vision are typically unsuitable for human vision.
Therefore, the development of scalable image coding methods that address these differences—allowing for optimized decoding for both human and machine use—is a promising area of research \cite{s7,s8,s9,s10,s11,s12,s13,s14,s15}. 
Such methods could effectively meet the diverse demands of human and machine image processing.

In scalable image coding, bridging the gap between human and machine vision is crucial.
Decoded images for machine vision are typically evaluated based on image recognition accuracy.
In contrast, decoded images for human perception are assessed using metrics like PSNR and SSIM, which measure the fidelity of pixel value reproduction, as well as LPIPS \cite{s16} and FID \cite{s17}, which correlate more closely with human subjective evaluation.
To address these dual objectives, scalable image coding methods have been developed to achieve high image recognition accuracy with faithful image reconstruction.
For human-oriented decoding, image quality is evaluated in terms of PSNR, aiming to reproduce the original image as closely as possible \cite{s18,s19,s20,s21,s22,t1}. 
However, this approach requires a substantial amount of information to reconstruct the original image, leading to a significant increase of data in the extension from machine vision to human vision.

In this paper, we propose a novel method to eliminate this additional information requirement, providing a seamless bridge between human and machine vision.
Drawing inspiration from the success of diffusion models in recent image compression techniques, we integrate diffusion models \cite{s23} into image coding frameworks for both human perception and image recognition models.
Specifically, the diffusion model, guided by the decoded image for machine vision, generates images suitable for human perception without the need for extra data.
Through experiments, we evaluate the bitrate and image quality of the generated images and demonstrate their effectiveness for human perception, offering a promising solution to unify the needs of both human and machine vision in scalable image coding.

\begin{figure}[bt]
   \centerline{\includegraphics[width=1.0\columnwidth]{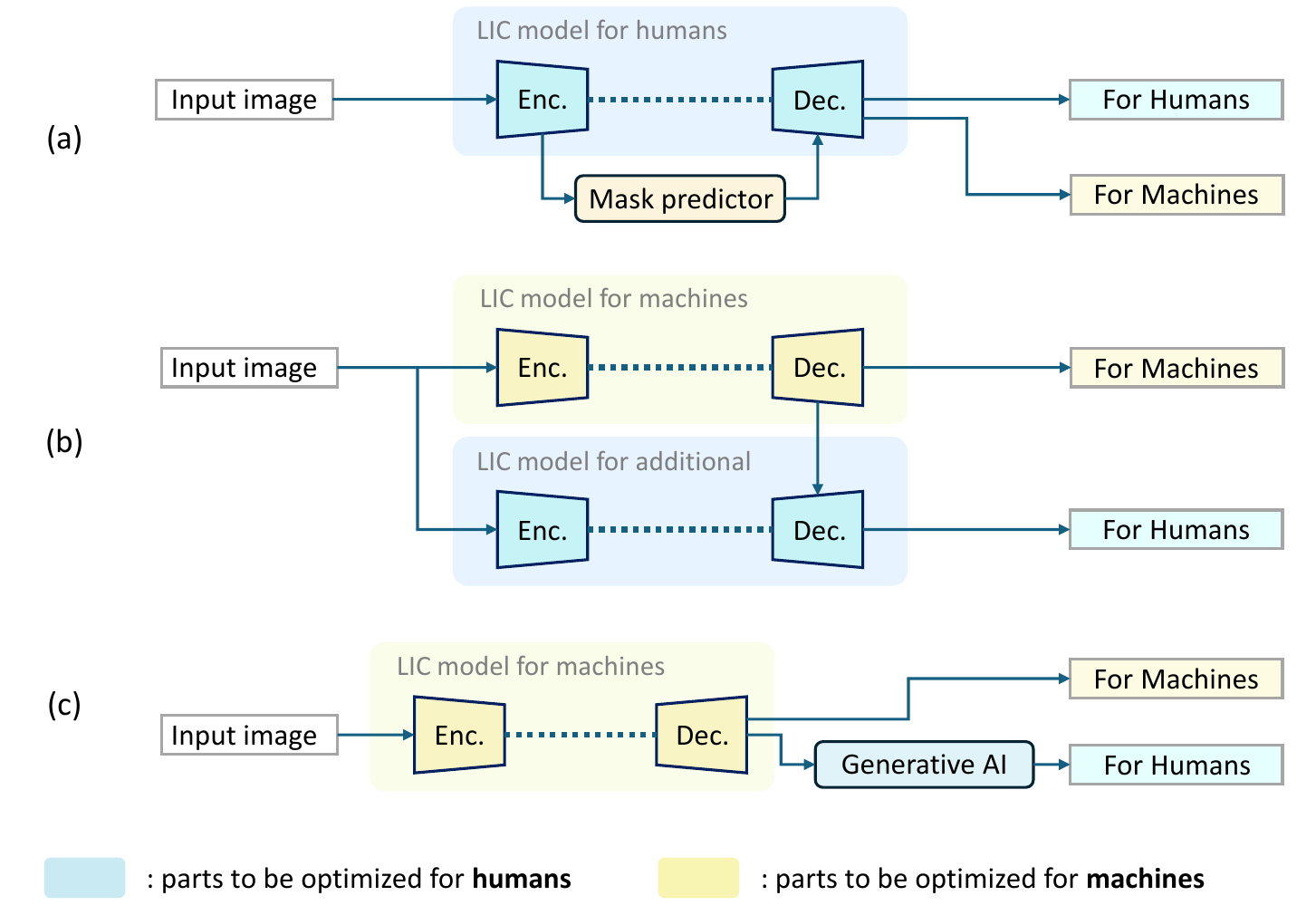}}
   \caption{Image processing in scalable image compression methods: (a) ICMH-Net, (b) ICMH-FF, (c) Ours.}
   \label{fig:process}
   \end{figure}

\begin{figure*}[bt]
   \centerline{\includegraphics[width=2\columnwidth]{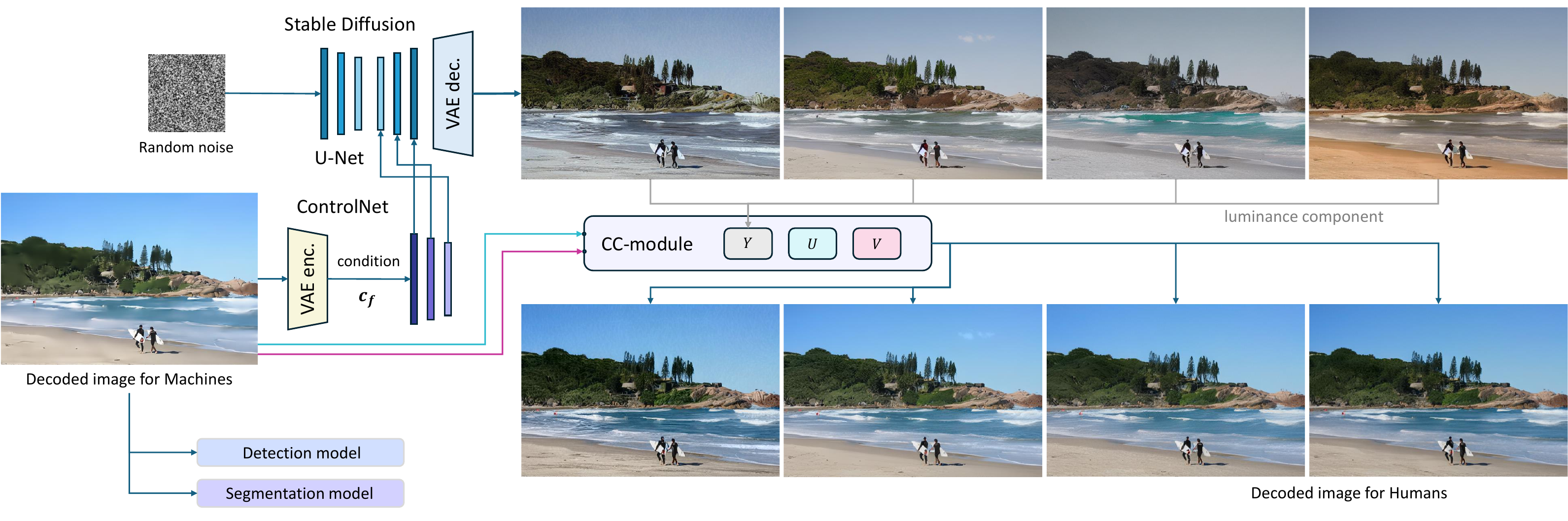}}
   \caption{Image processing flow in the proposed method. The ICM method serves as a condition to generate images optimized for human perception. The generated image is then input into the CC-module, which restores color elements approximating the original image.}
   \label{fig:flow}
   \end{figure*}
\begin{figure*}[bt]
   \centerline{\includegraphics[width=2\columnwidth]{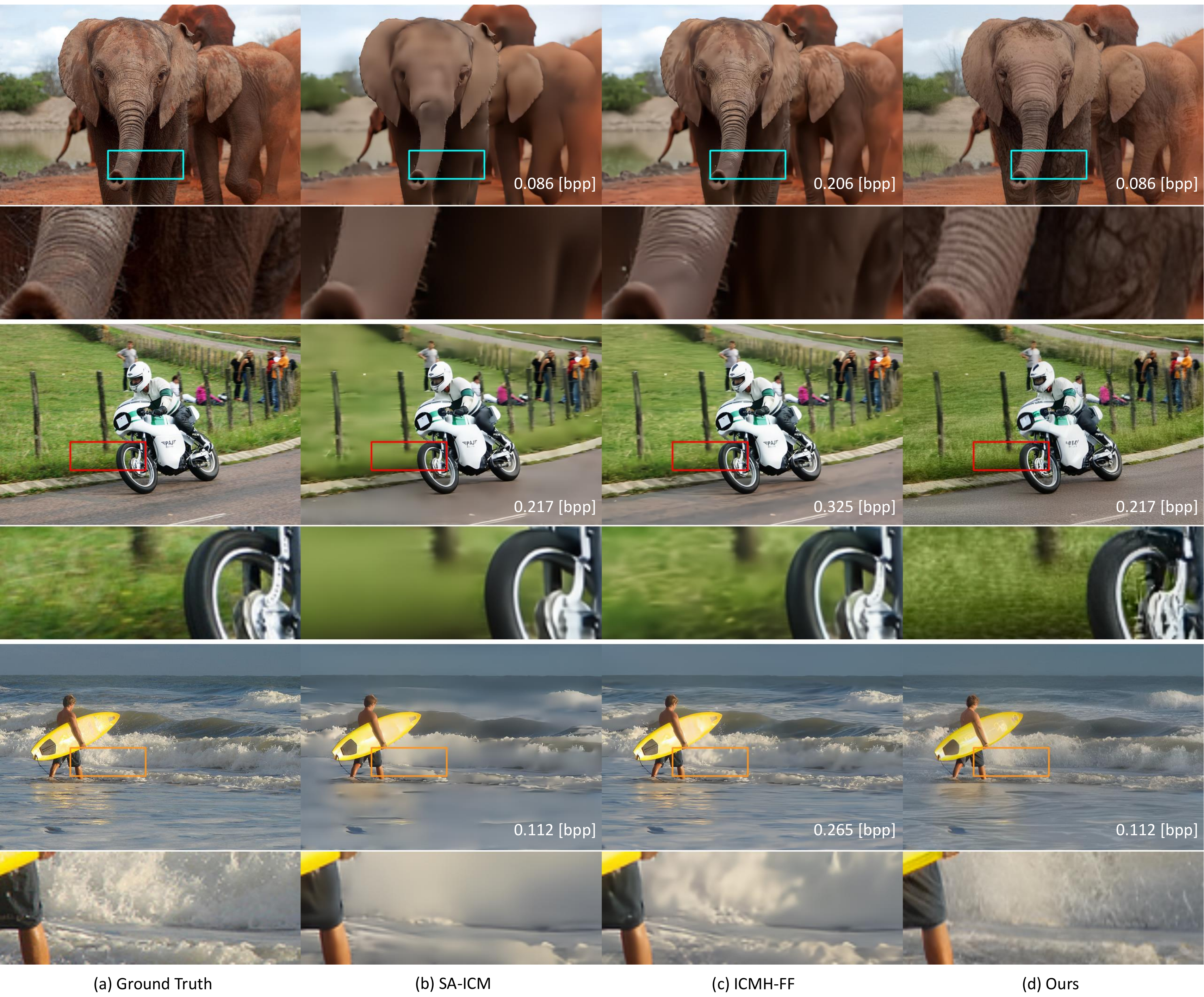}}
   \caption{Examples of original and decoded images: (a) Original image, (b) Decoded image for machine vision using SA-ICM, (c) Decoded image for human vision using ICMH-FF, (d) Decoded image for human perception using the proposed method.}
   \label{fig:coded}
   \end{figure*}

\begin{figure*}[t]
   \centerline{\includegraphics[width=2\columnwidth]{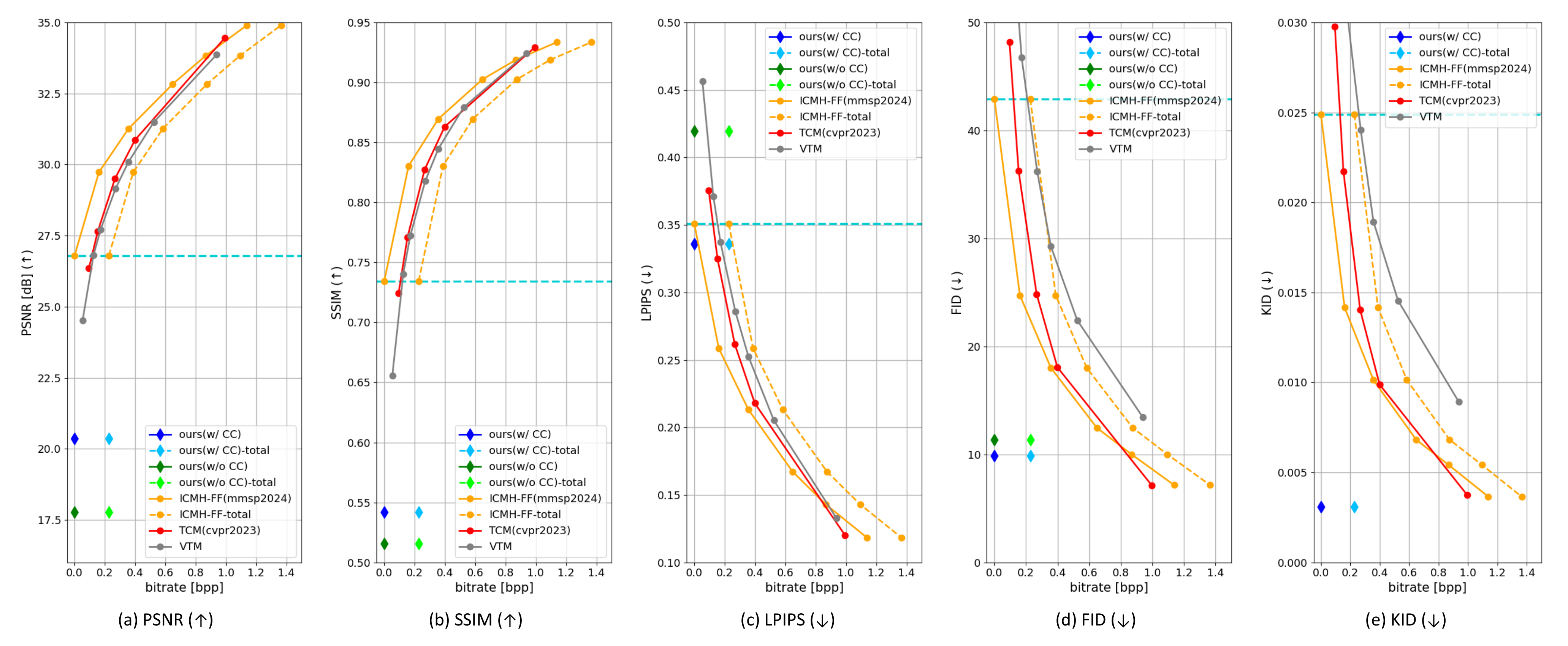}}
   \caption{Image compression performance of the proposed and comparative methods. Five metrics are used to evaluate image quality for human visual perception: (a) PSNR($\uparrow$), (b) SSIM($\uparrow$), (c) LPIPS($\downarrow$), (d) FID($\downarrow$), and (e) KID($\downarrow$).}
   \label{fig:per}
   \end{figure*}

\section{Related Works}
\subsection{Image Coding for Human Perception and Machine Vision}
There are two primary types of scalable image coding methods designed for both human and machine use.
The first approach builds on image compression methods optimized for human perception.
When decoding images for machine vision, this removes image information that is unnecessary for machines.
Examples of this approach include ICMH-Net \cite{s18} and Adapt-ICMH \cite{s19}.
These methods leverage Learned Image Compression (LIC) \cite{s24,s25,s26,s27}, a neural network-based image compression framework.
By incorporating components optimized for image recognition into a LIC model tailored for human perception, they enable efficient image compression for machine vision.
The image processing flow of ICMH-Net is illustrated in Figure \ref{fig:process}-(a).
A mask predictor is employed to identify which image features are necessary for image recognition and which are not.
By masking and removing the unnecessary features, the image can be decoded for machine vision using less information than is required for human perception.

The second approach uses image compression methods optimized for image recognition as a foundation and incorporates additional information when decoding images for human perception.
This framework includes ICM method and techniques for compressing supplementary information.
VVC+M is a scalable image coding method based on ICM model\cite{s22}.
In this method, the image is first reconstructed from the decoded features intended for machine vision.
This reconstructed image contains only the image information necessary for image recognition.
The difference between this reconstructed image and the original image is then compressed using the VVC-inter codec \cite{s28}, effectively treating the missing information for human perception as additional data.
ICMH-FF, on the other hand, combines an ICM method called SA-ICM \cite{s29,s30} with a LIC model for  compressing the additional information \cite{s20}.
The image processing flow of this method is shown in Figure \ref{fig:process}-(b).
In this figure, the top layer represents the image compression process for machine vision, while the bottom layer shows the process for compressing the additional information.
To evaluate the performance of image compression for machine vision, VVC+M uses an object detection model, while ICMH-FF utilizes both an object detection model and a segmentation model.
For evaluating image compression performance aimed at human perception, both methods use PSNR as the metric, focusing on faithfully reproducing the original image.

\subsection{Diffusion-based Image Compression}
Diffusion models \cite{s23} have gained significant attention as a new approach to image generation, offering a promising alternative to Generative Adversarial Networks \cite{s31}.
Among these, Stable Diffusion \cite{s32} has become widely adopted due to its user-friendly design.
By using text prompts or semantic maps as input conditions, the output image can be effectively controlled.
The introduction of ControlNet \cite{s33} has further advanced this capability by allowing the diffusion process to be guided with arbitrary conditions, making it even easier to generate desired images.
This progress in controlling the image generation process has spurred a growing interest in applying diffusion models to image compression tasks \cite{s34}.

PIC \cite{s35} is an image compression method that leverages a diffusion model to generate images, using a text prompt as the condition.
The bitrate required to decode the image corresponds to the amount of information contained in the text prompt.
PICS \cite{s35} expands on this approach by incorporating sketch images alongside text to control the diffusion process.
The spatial constraints introduced by the sketch images enable precise control over the shapes and positions of objects and backgrounds.
In PICS, the image bitrate is determined by the combined information in the text prompt and sketch image.
Additionally, other studies have proposed methods for controlling the generated image's characteristics, such as using a color map to define the color scheme or low resolution image  of the original image to guide the generation process.

Diffusion-based image compression methods are not well-suited for faithfully reproducing the original image and have been reported to perform worse than LIC and traditional rule-based codecs like JPEG, HEVC \cite{s36}, and VVC \cite{s28} in metrics such as PSNR and SSIM.
However, numerous studies have demonstrated that these methods outperform existing image compression techniques in evaluation indices like LPIPS and FID, which are believed to align more closely with human subjective perception.
Therefore, diffusion-based image coding methods are expected to be an image compression method for human perception.

\section{Proposed Method}
\subsection{Generative processes conditioned on ICM method.}
We use the decoded image designed for machine vision as a condition to generate an image suitable for human vision.
We adopt SA-ICM \cite{s29} as the image compression method for machine vision, which is an image compression method for object detection and segmentation models.
This method decodes the contours of objects within an image while discarding other textures.
The resulting decoded image retains the position and size of objects and the background but lacks detailed textures.
An example of such a decoded image is shown in Figure \ref{fig:coded}-(b).
In the decoded image intended for machines, for instance, several elephants are recognizable in the center of the image, but its detailed patterns are not visible.

To restore the textures lost during compression, we leverage generative AI.
The proposed image processing flow is shown in Figure \ref{fig:flow}.
The generative AI in our proposed method combines Stable Diffusion \cite{s32} and ControlNet \cite{s33}.
Stable Diffusion serves as the diffusion model, while ControlNet is used to guide the generative process.
The ControlNet is fed with the SA-ICM decoded image and trained to regain the texture lost due to compression.
The loss function for training ControlNet is defined as follows:
\begin{equation}
   \mathcal{L}=\mathbb{E}_{z_{0},\mathbf{t},\mathbf{c_{t}},\mathbf{c_{f}},\epsilon\sim\mathcal{N}(0,1)}\lbrack\lVert \epsilon-\epsilon_{\theta}(z_{t},\mathbf{t},\mathbf{c_{t}},\mathbf{c_{f}}) \lVert_{2}^{2}\rbrack.
\end{equation}
In (1), $\mathbf{t}$ represents the time step, $z_{0}$ denotes the original latent, and $z_{t}$ represents the latent after adding noise at step $\mathbf{t}$. 
$\mathbf{c_{t}}$ and $\mathbf{c_{f}}$ refer to the text prompt and image condition, respectively.
In the training process of the proposed method, $\mathbf{c_{t}}$ is an empty string, and $\mathbf{c_{f}}$ is derived from the features of the decoded image from SA-ICM. 
By predicting the noise present in the latent, it becomes possible to generate images starting from random noise.
Training ControlNet to transform images designed for recognition models into images suitable for human perception allows seamless extension from machine vision to human vision without relying on additional information.
The only input needed to generate images for human perception is the decoded image intended for image recognition.
Using this decoded image as a guide, the image for human visual perception is generated from random noise through the diffusion model.
In other words, the information required to decode an image for humans is identical to the amount needed to decode an image for machine vision.

\subsection{Color control of generated images.}
The images generated for human visual perception, conditioned on the ICM method, successfully reproduce the positions and locations of objects and backgrounds in the images. 
However, they are not well-suited for accurate color reproduction.
In the SA-ICM decoded images, much of the original texture is lost. To make these images suitable for human use, they need to be enhanced with various colors to restore texture. 
In other words, the generative AI adds textures in diverse colors based on the SA-ICM decoded images. 
This process, however, can result in colors that differ from the original image, leading to inconsistencies in color reproduction.

To address this, we apply a Color Controller (CC) module to the generated images. The CC-module adjusts the color components of the images produced by the diffusion model. 
Specifically, the color components in the generated image are replaced with those from the SA-ICM decoded image, while the luminance component is preserved from the generative AI output. 
This approach helps the generated image achieve colors that closely match the original.

\section{Experiment}
\subsection{Experimental Details and Evaluation Methods.}
We conducted experiments to evaluate the effectiveness of the proposed method in extending machine vision to human vision.
First, we trained a generative AI model to bridge the gap between these two types of vision.
For this training process, we prepare decoded images from the ICM method along with their corresponding original images.
To obtain these decoded images, we utilize the official implementation of SA-ICM \cite{s29} for image decoding, tailored for image recognition tasks.
An example of a decoded image is presented in Figure \ref{fig:coded}-(b).
These machine-oriented decoded images serve as the input for ControlNet, which is trained to reproduce the original images.
The loss function for this training is defined in (1).
The COCO dataset \cite{s38} serve as the benchmark for our experiments, utilizing 118,287 COCO-train images for training and 5,000 COCO-val images for testing.
The text prompt $\mathbf{c_{t}}$ is left as an empty string during both the training and testing phases.

In this experiment, the proposed method and comparative approaches are primarily evaluated based on bitrate and image quality for human perception.
The comparative method, ICMH-FF, is a scalable image coding technique designed for both humans and machines \cite{s20}.
To ensure a fair comparison, the same ICM method is employed in the implementation of both ICMH-FF and the proposed method.
Additionally, common image compression methods for human perception, such as VVC \cite{s28} and TCM \cite{s40}, are included as benchmarks. VVC is a rule-based video compression standard, while TCM is a LIC model that represented the state of the art in 2023.
The evaluation of image quality is conducted using metrics such as PSNR, SSIM, LPIPS \cite{s16}, FID \cite{s17}, and KID \cite{s39}.

\subsection{Experimental Results.}
Figures \ref{fig:coded}-(c) and \ref{fig:coded}-(d) showcase examples of decoded images for human perception using the comparative and proposed methods, respectively.
The bitrate required to decode each image is also indicated in the figure.
Qualitative evaluation reveals that the proposed method successfully recovers textures lost during the compression process of the ICM method.
Unlike conventional approaches, our method effectively depicts fine details, such as elephant's wrinkles and wave splashes, even without relying on additional information.

For the quantitative evaluation, we analyzed the relationship between five image quality indicators and the bitrate, as illustrated in Figures 4-(a) to 4-(e).
The blue and green points represent the compression performance of the proposed method with and without the CC-module, respectively.
These bitrates indicate only the amount of information needed for transitioning from machine vision to human vision, which remains at 0 [bpp] in all graphs.
The light blue and lime green dots represent the total amount of information required by the proposed method to decode images for human perception, including the information necessary for decoding images suitable for machines.
The orange curve represents the compression performance of ICMH-FF, where the solid line indicates the amount of additional information, and the dotted line reflects the total of this additional information and the bitrate of the image for machines.
The red and gray curves show the compression performance of TCM and VTM, respectively.
Figures \ref{fig:per}-(a) and \ref{fig:per}-(b) reveal that the proposed method is less effective than conventional methods in terms of accurately reproducing the original image's pixel values, as measured by distortion metrics.
However, Figures \ref{fig:per}-(d) and \ref{fig:per}-(e) demonstrate that our method significantly surpasses conventional methods in perceptual metrics like FID and KID, which are closely aligned with human subjective evaluations.
These findings indicate that while the proposed vision enhancement method may not excel in fidelity-based decoding, it is highly effective at generating perceptually meaningful images for human interpretation.

Table \ref{tab:comp-human} compares the bitrates of the methods at equivalent LPIPS and FID scores, while Table \ref{tab:comp-machine} presents the compression performance of each method in image compression for machine vision.
For the image recognition model, Yolov5 \cite{s41} is used for the object detection and Mask R-CNN \cite{s42} for the segmentation.
In Table \ref{tab:comp-human}, the bit rates for both the proposed method and ICMH-FF reflect the total bitrate, which includes the information required for machine decoding.
In Table \ref{tab:comp-machine}, since both methods utilize the same ICM method, their image recognition performance is identical.
These comparisons highlight that the proposed method, when combined with the ICM method, forms an effective approach to scalable image compression for both human and machine vision.
Notably, the proposed method enables the transformation of decoded images for machines into human-perceptible images with no additional information, serving as a valuable tool to bridge the gap between these two types of visual processing.

\begin{table}[t]
\centering
\caption{Comparison of image compression performance for humans between the proposed and conventional methods.} \label{tab:comp-human}
\small
\begin{tabular}{c|cc}
   \hline
   method & LPIPS & FID\\
   \hline
   \hline
   TCM \cite{s40} & 0.137 ($\pm$0.0[\%]) & 0.710 ($\pm$0.0[\%])\\
   \hline
   ICMH-FF \cite{s20} & 0.235 (+71.53[\%]) & 1.016 (+43.10[\%])\\
   \hline
   Ours & 0.227 (+65.69[\%]) & 0.227 (-68.03[\%])\\
   \hline
\end{tabular}
\end{table}

\begin{table}[t]
   \centering
   \caption{Comparison of image compression performance for machines between the proposed and conventional methods.} \label{tab:comp-machine}
   \small
   \begin{tabular}{c|cc}
     \hline
     method & detection \cite{s41} & segmentation \cite{s42}\\
     \hline
     \hline
     TCM \cite{s40}  & 0.356 ($\pm$0.0[\%]) & 0.342 ($\pm$0.0[\%])\\
     \hline
     ICMH-FF \cite{s20} & 0.227 (-36.23[\%]) & 0.227 (-36.23[\%])\\
     \hline
     Ours & 0.227 (-36.23[\%]) & 0.227 (-36.23[\%])\\
     \hline
   \end{tabular}
 \end{table}

\section{Conclusion}
In this paper, we propose a novel method for converting images for machine vision into images for human visual perception.
Our experiments demonstrated that by guiding the generation process of generative AI under the constraints of the ICM method, it is possible to create images suitable for humans from machine-targeted images.
This approach serves as a bridge, seamlessly connecting machine vision and human perception.
Future research should explore applying the proposed vision extension method to other ICM techniques to further validate its effectiveness.



\end{document}